\documentclass[journal]{IEEEtran}
\usepackage{multirow}
\usepackage[skip=10pt]{caption}
\usepackage{booktabs}

\usepackage{times}
\usepackage{epsfig}
\usepackage{graphicx}
\usepackage{amsmath}
\usepackage{amssymb}
\usepackage{color}
\usepackage[normalem]{ulem}
\usepackage{cite}

\ifCLASSINFOpdf
\else
\fi
\usepackage[caption=false,font=footnotesize]{subfig}
\usepackage{url}

\usepackage{hyperref}

\begin{document}

%
\title{Video Classification With CNNs: Using The Codec As A Spatio-Temporal Activity Sensor}
%
%
%

\author{Aaron Chadha, Alhabib Abbas and Yiannis Andreopoulos,~\IEEEmembership{Senior Member,~IEEE}
\thanks{AC and AA are with the Electronic and Electrical Engineering Department, University College London, Roberts Building, Torrington Place, London, WC1E 7JE, UK (e-mail:
\{aaron.chadha.14, alhabib.abbas.13\}@ucl.ac.uk). YA is with the Electronic and Electrical Engineering Department, University College London, Roberts Building, Torrington Place, London, WC1E 7JE, UK, and also with Dithen, 843 Finchley Road, London, NW11 8NA, UK, \href{www.dithen.com\#dithen}{www.dithen.com} (e-mail:\ i.andreopoulos@dithen.com). We acknowledge support from: EPSRC (project EP/P02243X/1), Innovate UK (project DELVE-VIDEO 132739), The Leverhulme Trust (RAEng/Leverhulme Senior Research Fellowship of Y. Andreopoulos) and the Royal Commission for the Exhibition of 1851 (Fellowship of A. Chadha). This work has been presented in part at the 2017 IEEE International Conference on Image Processing, Beijing, China. }}

%
%

\markboth{IEEE Transactions on Circuits and Systems for Video Technology, to appear}%
{Shell \MakeLowercase{\textit{et al.}}: Bare Demo of IEEEtran.cls for IEEE Journals}
%



\maketitle

\begin{abstract}
We investigate video classification via a two-stream convolutional neural network (CNN) design that directly ingests information extracted from compressed video bitstreams. Our approach begins with the observation that all modern video codecs divide the input frames into macroblocks (MBs). We  demonstrate that selective access to MB motion vector (MV) information within compressed video bitstreams can also provide for selective, motion-adaptive, MB pixel decoding (a.k.a., MB texture decoding). This in turn allows for the derivation of spatio-temporal video activity regions at extremely high speed in comparison to  conventional full-frame decoding  followed by optical flow estimation.  In order to evaluate the accuracy of a video classification framework based on such activity data, we independently train two   CNN  architectures on MB  texture and MV correspondences and then fuse their scores to derive the final classification of each test video. Evaluation on two standard datasets shows that the proposed approach is  competitive to the  best two-stream video classification approaches found in the literature. At the same time: \textit{(i)} a CPU-based realization of our MV extraction is over 977 times faster than GPU-based optical flow methods; \textit{(ii)} selective  decoding   is up to 12 times faster than full-frame decoding;  \textit{(iii)} our proposed spatial and temporal CNNs  perform inference at 5 to 49  times lower cloud computing cost than the fastest methods from the literature.
\end{abstract}


%
\IEEEpeerreviewmaketitle

\section{Introduction}
%
%
%
%
\IEEEPARstart{F}{or the} last 50 years, the holy grail of machine learning with visual data has been to translate \textit{pixels to concepts} \cite{lecun2015deep}, e.g., classify a pixel-domain video sequence according to its contents (�tennis match�, �horror film�, �cooking show�, �people marching,...��). However, it has been argued recently \cite{posch2015giving,tan2015benchmarking} that that there is no strong scientific basis for this focus on pixels: it simply stems from the 140-year old legacy of video being displayed as sequences of still frames comprising the raster-scan of picture elements. Pixel-domain video representations are in fact known to be cumbersome for cognitive video analysis,
primarily due to two aspects: \textit{(i)} all state-of-the-art methods for
high-level semantic description in video require memory- and compute-intensive decoding and
 pixel-domain processing, such as optical flow
calculations \cite{simonyan2014two,varol2016long,zhang2016real}; \textit{(ii)}  the
high resolution \& high frame-rate nature of decoded video and the format
inflation (from standard to super-high definition, 3D, multiview, etc.) require
highly-complex convolutional neural networks (CNNs) that impose massive computation and storage
requirements \cite{szegedy2015going}. 

{Inspired by these observations, hardware designs of neuromorphic sensors, a.k.a., silicon retinas  \cite{posch2015giving,tan2015benchmarking}, have been proposed recently. Unlike conventional active pixel sensors, silicon retinas create a spatio-temporal activity representation (a.k.a., spatio-temporal spike event stream) representing the illumination changes caused by object motion, while incurring very low power and low latency.} Nevertheless, due to the need to be compliant to display technology,  pixels and video frames are here to stay: after all, pixel-based video frames are being used today within all conversational, entertainment and mainstream visual surveillance services. However, because of storage and data-transfer limitations, all camera chipsets and video processing pipelines provide compressed-domain video formats like MPEG/ITU-T AVC/H.264 \cite{wiegand2003overview} and HEVC \cite{sullivan2012overview}, or open-source video formats like VP9 \cite{mukherjee2013latest} and AOMedia Video 1 instead of uncompressed (pixel-domain) video. Alas, the state-of-the-art in CNN-based classification and recognition in video \cite{simonyan2014two,varol2016long,zhang2016real} ignores the fact that  video codecs can be tuned at the macroblock (MB) level. For example, the MPEG/ITU-T AVC/H.264 and HEVC codecs divide the input video frames into $16 \times 16$ pixel MBs that form the basis for the adaptive inter (and intra) prediction. Inter-predicted MBs are (optionally) further partitioned into blocks that are predicted via motion vectors (MVs) that represent the displacement from matching blocks in previous or subsequent frames. 

The research hypothesis of this paper is to {consider  the video encoder as an imperfect-yet-highly-efficient �``sensor''� that derives spatio-temporal activity representations with minimal processing}. With regards to the temporal activity, we demonstrate that we can obtain MV representations from the compressed bitstream that are highly correlated with optical flow estimates. We then propose a three-dimensional CNN that directly
leverages on such MB MV information and compensates for the sparsity of
these MB MVs with larger temporal extents. With regards to the spatial activity, we show that selective MB texture decoding can take place based on thresholding of the MB MV information. By superimposing such selectively-decoded MB texture information on sparsely-decoded frames, we obtain spatial representations that are shown to be visually similar to the corresponding fully-decoded  video frames. This allows for the parsimonious use of a spatial CNN to augment the classification results derived from the temporal stream. Our results with the fusion  of this two-stream CNN design on two widely-used datasets show that competitive accuracy is  obtained against the state-of-the-art, with extraction and classification complexity that is found to be one to three orders-of-magnitude lower than that of all previous approaches based on pixel-domain video. Importantly, the complexity gains from using compressed MB bitstreams are future-proof: as video, multi-view and 3D video format resolutions and frame rates increase to accommodate advances in display technologies, the gains provided by such approaches will increase commensurably to the change in spatio-temporal sampling. This paves the way for exabyte-scale video datasets to be
newly-discovered and analysed over commodity hardware.

\section{Related Work } \label{sec:related_work}

Due to their outstanding performance in image classification \cite{krizhevsky2012imagenet}, deep convolutional neural networks (CNNs) have
recently come to the forefront in video classification, remaining competitive to or surpassing performance of traditional methods derived on hand-crafted features, such as improved dense trajectories (IDT) \cite{wang2011action, wang2013action}.  With increasing dataset sizes and complexity in classification and 
retrieval,   deep and scalable  CNN architectures are capable of  learning more
complex representations. Karpathy \textit{et al.} \cite{karpathy2014large} proposed  extending the CNN architecture from image to video by performing spatio-temporal convolutions in the first convolutional layers over a 4D video chunk $V \in \mathbb{R}^{W \times H \times K\times  T}$, where $W,H$ are the spatial dimensions, $K$ is the number of channels and $T$ is the number of frames in the chunk.  This is the premise behind their slow-fusion architecture, which uses 3D convolutions on RGB frame chunks in the first 3 layers, thus encompassing the full spatio-temporal extent of the input. Tran \textit{et al.} \cite{tran2015learning} attempted to improve accuracy by using a deep �3D CNN architecture (resembling VGGnet \cite{simonyan2014very}) together with spatio-temporal convolutions and pooling in all layers,  albeit with heavy computational cost. In this paper, we show that a deeper 2D CNN architecture ingesting a single RGB\ frame should be sufficient to perform competitively to 3D architectures, whilst providing simplicity in training and pre-processing the inputs.     

 Indeed, Simonyan \textit{et al.} \cite{simonyan2014two}, argue that the problem is not the depth or spatio-temporal extent of the architecture but rather the nature of the RGB input that does not effectively present motion information to the CNN.  They propose using a 2D architecture with dense optical flow to represent  the temporal component of the video.  Notably, this temporal CNN is shown to outperform an equivalent 2D spatial stream ingesting RGB frames.  Performance can be improved further by fusing the temporal and spatial streams using a simple score averaging.  This two stream architecture achieves 88.0\% on UCF-101. Nevertheless, the computational cost remains high due to the requirement to extract Brox optical flow \cite{brox} for the temporal stream.
In this paper, we also employ a two stream  architecture to model the temporal motion and scene information independently. In order to reduce the computational overhead from having to fully decode and process the video, we circumvent the highly-complex optical flow calculation by using  MV representations extracted directly from the video codec. 

Recent work \cite{zhao2016real} has used hand-crafted features, in the form of   a spatio-temporal Bag-of-Words approach on refined MV representations for object-based segmentation in video. The use of codec MV representations has also been proposed for action recognition by Kantorov and Laptev \cite{mpegflow}. However, their approach uses Fisher vectors, which achieves lower accuracy in standard action recognition datasets. {Recently, Zhang \textit{et al.} \cite{zhang2016real}  utilized codec MVs as an input to a 2D CNN in their action recognition, termed EMV-CNN, but their framework requires optical-flow based training (in their proposed teacher net). This constrains the training to relatively small volumes of video content and requires upsampled P and B-frame MV fields during inference due to the teacher-student supervision transfer \cite{zhang2016real}. 
Our paper is the first to achieve state-of-the-art performance \textit{without} the use of optical flow and with substantially higher speed in comparison to EMV-CNN and all existing alternatives. 
 In addition,  given that  the spatial stream predominantly learns on scene information that tends to be persistent  across frames, we gain by  sparse frame decoding combined with motion adaptive  superpositioning of  decoded MB texture information to generate   intermediate frames at a finer temporal resolution\footnote{While initial results with the proposed MV-based CNN approach have been presented in our corresponding conference paper \cite{chadha2017},  MV-based selective MB texture decoding and the fusion of the temporal 3D-CNN with a spatial CNN are proposed here for the first time.}}.  

One of the main issues with the related work described above is the short temporal extent of the inputs \cite{xu2017two,zhao2017pooling}; each input is a small group of frames that only encapsulates a second or so of the video.  This does not account for cases where temporal dependencies extend over longer durations.  {Feichtenhofer \textit{et al.} \cite{feichtenhofer2016convolutional} attempted to resolve this issue by using multiple copies of their two stream network. The copies are spread over a coarse temporal scale, thus encompassing both coarse and fine motion information with an optical flow input.  The  architecture is spatially and then temporally fused   using a 3D convolution and pooling. Despite achieving state-of-the-art results on UCF-101 and HMDB-51 datasets, this approach  requires very computationally demanding processing for both training and testing.  This is in part due to the joint training, which can cause a parameter explosion depending on the chosen architecture and fusion method (see Table 1 of Feichtenhofer \textit{et al.} \cite{feichtenhofer2016convolutional}).}  Alternatively, Laptev \textit{et al.} \cite{varol2016long} argue that increasing the temporal extent is simply a case of taking the optical flow component  over a larger temporal extent.  In order to minimize the complexity of the network, they downsize the frames,  thus reducing the spatial dimensions.  Combining their two stream architecture with improved dense trajectories yields 92.7\% on  UCF-101. {Finally, Action VLAD \cite{girdhar2017actionvlad} was proposed as the means to encompass a longer temporal extent by pooling features spatio-temporally into a VLAD descriptor.  However,  this aggregation requires \textit{(i)} additional VLAD pre-processing computation and \textit{(ii)} multiple CNN copies during training in order to cover the entire temporal extent of the video. These requirements raise significant demands in GPU  memory  and  CPU  preprocessing  capability.}
Contrary to these proposals, our  temporal stream input (as extracted from the video codec), is inherently of low spatial resolution, thus allowing for significantly lower complexity in processing. This enables us to feed an even larger temporal extent into our 3D CNN.  We also improve the architecture by minimizing the number of activations in the lower layers with a temporal downsizing (using a stride of 2); this allows us to reduce the bottleneck in processing the temporal input volume.

\begin{figure}[t!] \centering \includegraphics[scale=0.4]{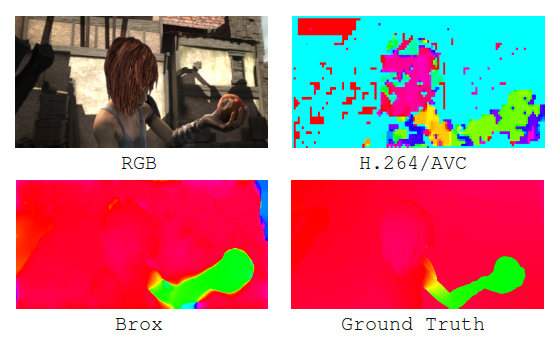}
        \caption{\label{fig:compmv} RGB\ frame from the MPI-Sintel dataset and pseudocolored images of the  motion information amplitude. The H.264/AVC MB motion vectors are correlated with Brox optical flow 
extracted from decoded video frames \cite{brox}\cite{ilg2016flownet}  and the ground-truth motion available for this synthetic video.} \end{figure}

Another method of generating CNNs with a long temporal extent is to integrate a recursive neural network (RNN) into the architecture.  In principle, an RNN provides for infinite temporal context up to the present frame.  Donahue \textit{et al.} \cite{donahue2015long}  use a 2D CNN to extract features from individual frames. These are subsequently fed into a stack of long short term memory (LSTM) networks  for sequence learning over the input.  Due to parameter sharing over time, this model scales to arbitrary sequence lengths.    Ng \textit{et al.} \cite{yue2015beyond} extend this  by  considering the effects of appending the CNN with feature pooling versus an LSTM, prior to class fusion. Their results demonstrate that pooling is a good alternative to using an LSTM and achieves competitive accuracy (88.2\%\ vs 88.6\% on UCF-101).
  They also note that simply appending a 2D CNN with an LSTM\ stack has its limitations.  For one, the LSTM\ is likely to only focus on global temporal motion, such as shot detection and not the fine temporal cues inherent in groups of consecutive frames.  In this paper, we decide against integrating  LSTMs into our framework as our temporal extent is sufficiently large to encompass the entire video duration.  \   

\begin{figure*}[ht] \centering \includegraphics[scale=0.3]{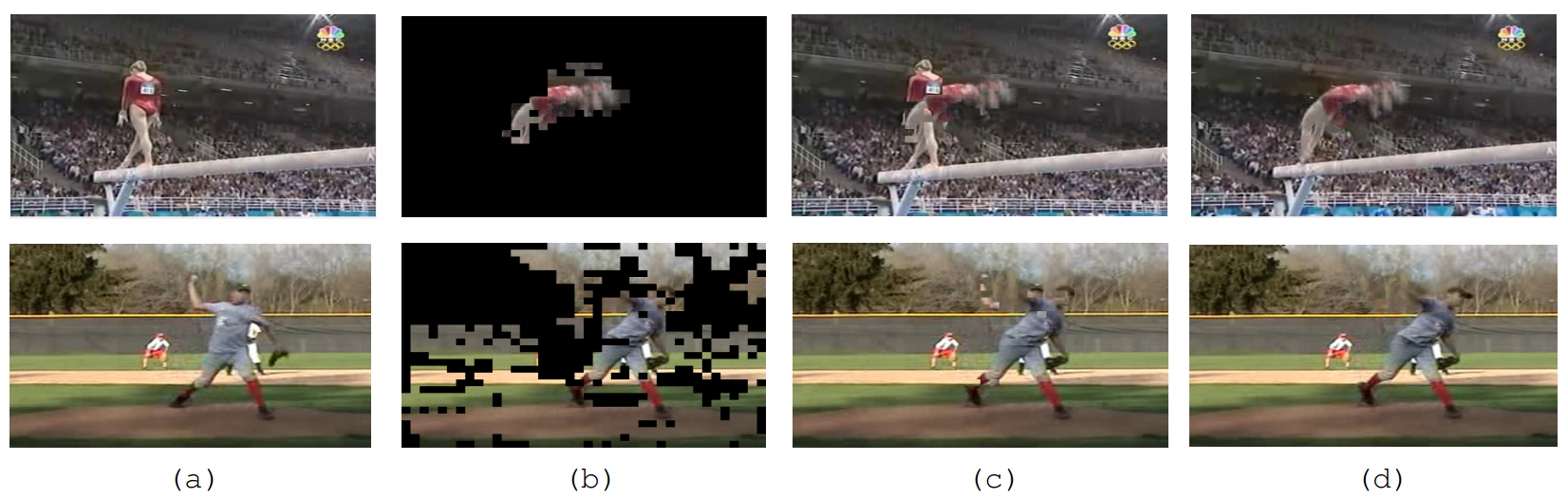}
        \caption{\label{fig:mvx_seldec} Two scenes from  UCF-101   with  \&\ without camera motion (top \&\ bottom row respectively); (a) Reference frame; (b) Selective decoding of MB texture ($A =0$); (c) Rendered frame; (d) Fully decoded frame corresponding to the rendered frame. } \end{figure*}

\section{Selective MB Motion Vector Decoding}
\label{sec:motion_estimation}

Video compression standards like MPEG/ITU-T AVC/H.264, and HEVC rely on motion estimation and compensation as their main method to decorrelate successive input frames. Macroblock motion vectors are
derived by temporal block matching and can be interpreted as
approximations of the underlying optical flow 
\cite{coimbradavies}\cite{mpegflow}, as shown in the example of Fig. \ref{fig:compmv}.

To derive a temporal activity map from encoded motion compensation parameters, we apply the
following steps:

\begin{enumerate}

\item Motion vectors are  extracted from certain compressed MB information of the utilized video codec\footnote{Based on FFMPEG's widely-used \texttt{libavcodec} library (which supports most MPEG/ITU-T standards) \cite{libavcodec}, {we make use of the \texttt{AVMotionVector} structure (declared within the \texttt{avutil.h} header file) as explained in the following. When \texttt{libavcodec}  attempts to read the compressed bitstream of a video frame using the \texttt{av\_read\_frame()} function, our MB MV extractor calls the \texttt{av\_frame\_get\_side\_data()} function to extract the MV parameters and place them in the \texttt{AVMotionVector} structure. Once the file parsing is completed, the horizontal and vertical coordinates of MVs of each MB within this structure are written in 16-bit integer binary format to disk in order to be used by the proposed 3D  CNN. By limiting the bitstream access to solely using this function for the MB MVs, one can achieve the speed gains reported in Section \ref{sec:Evaluation}.}}.

\item If necessary, motion vectors are interpolated spatially to generate a finer representation of motion activity in the video, {i.e., with resolution corresponding to $8\times 8$ or $4\times 4$ blocks, and also to ``fill in'' for macroblocks where the video encoder may have used an intra prediction mode}.  
\end{enumerate}

For the spatial stream, we employ selectively-decoded MB texture information using the extracted MVs as activity indicators. We do this by  decoding one frame every $X$ frames, with $X$ set  to $\inf,$ indicating that only the first frame of the video is decoded. In between fully-decoded frames, ``rendered'' frames can be produced at frame interval $R$, with $1 \leq R \leq X$. Each rendered frame is initialized as a copy of the immediately preceding fully-decoded frame. Then, texture information  at  active MB positions is decoded and replaces the initialized values in the corresponding locations in the rendered frame. Two examples of this process are shown in Fig. \ref{fig:mvx_seldec}. We consider the area within a macroblock to be active when the corresponding MV information exceeds a specified threshold $A$, $A \geq 0$. As an illustration, Fig. \ref{fig:activity_region} shows a grayscale activity map derived from the MVs  of Fig. \ref{fig:mvx_seldec}(b). {To achieve such  blockwise selective MB texture decoding via the  \texttt{libavcodec} library \cite{libavcodec}, we use the  motion vectors from \texttt{AVMotionVector} to access \texttt{AVFrame::data} and write decoded MB texture data wherever the conditions specified by $\{X,R,A\}$ are met.} By increasing the values for $\{X,R,A\}$  we can decrease the frequency of full decoding and selective MB texture decoding in order to achieve any extraction speed desired within a practical application context. In addition, even though it is not explored in this paper, we can investigate adaptive control of $\{X,R,A\}$ based on the average MV activity level within each video sequence.

\section{Proposed Framework For Compressed-domain Classification}
\label{sec:network_design}

In this section we describe the proposed framework for training a temporal stream of MB motion vectors extracted directly from the video bitstream and a spatial stream comprising selective (motion-dependent) MB RGB texture decoding, and consider how the two streams can be fused during testing.  \    \subsection{Network Input} \label{sec:network_input}   
\subsubsection{Temporal Stream}

For our temporal stream input, we extract and retain only P-type  MB MVs, i.e., uni-directionally predicted MBs \cite{sullivan2012overview,wiegand2003overview}.  The standard  UCF-101
\cite{soomro2012ucf101} and HMDB-51 \cite{kuehne2011hmdb} datasets are composed of   $320 \times 240$ RGB pixels per frame. Therefore, for a frame comprising P-type MBs, a block size
of $8 \times 8$ pixels results in a motion vector field $\boldsymbol{\Phi}_\text{T} \in
\mathbb{R}^{W_\text{T} \times H_\text{T} \times K_\text{T}}$ of dimension $40 \times 30 \times 2$, where $W_\text{T}
\times H_\text{T}$ is the motion vector spatial resolution and  the number of channels $K_\text{T}=2$ refers to  the $\delta x$ and $\delta y$ motion vector components.

In order to compensate for the low spatial resolution $W_\text{T} \times H_\text{T}$, we take
a long temporal extent  of motion vectors over $T > 100$ consecutive P frames.
This is contrary to recent proposals based on high-resolution optical flow
\cite{simonyan2014two, karpathy2014large}, which typically ingest only a few 
frames per input (typically around 10). This is because, even with the latest GPU hardware, a long temporal extent cannot be processed without sacrificing the spatial resolution
of the optical flow 
\cite{simonyan2014two, karpathy2014large}. On the other hand, given that our MB motion vector input is inherently low-resolution, it is amenable to a longer temporal extent, which is more likely
to include the entirety of relevant action that is essential for the correct classification of the video. For example, we have found that the accuracy increases greatly for UCF-101 evaluated
on our 3D CNN when moving from 10 to 100 frames, but eventually plateaus when
$T$ becomes sufficiently large such that the input extends to almost all
P-type frames of the video files of the dataset.  Therefore, we fix the temporal extent
$T$ to 160, which is roughly the average number of P-frames per video in UCF-101.

\begin{figure}[ht] \centering \includegraphics[scale=0.3]{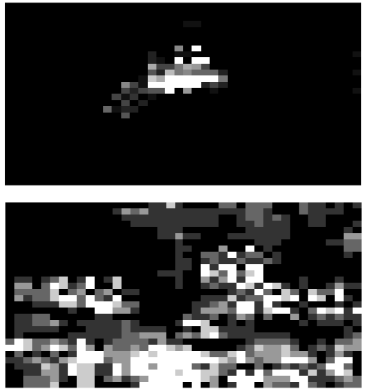}
        \caption{\label{fig:activity_region} MV activity maps  corresponding to Fig. \ref{fig:mvx_seldec}(b).} \end{figure}

In order to make our network input independent of the video resolution, we use a
fixed spatial size  $N_\text{T} \times N_\text{T}$ which is cropped/resized  from
$\boldsymbol{\Phi}_\text{T}$. {In this paper, we set $N_\text{T} = 24$, which is large enough to encompass the action region without compromising accuracy, whilst allowing for data augmentation via random cropping.  Furthermore, given it is divisible by eight, the P-frames can be continuously downsampled by a factor of 2  with pooling layers, without requiring any padding.}   Our final network input
$\boldsymbol{\hat\Phi}_\text{T} \in \mathbb{R}^{N_\text{T} \times N_\text{T} \times K_\text{T} \times T}$ is thus 4D
and can be ingested by a 3D CNN. As exemplified in numerous works
\cite{karpathy2014large, tran2015learning}, the advantage of using a 3D CNN
architecture with a 4D input, versus stacking the frames as channels and using a
3D input of size ${N_\text{T} \times N_\text{T} \times TK_\text{T}}$ with a 2D CNN, is that, rather than collapsing to a 2D slice  when convolving within the CNN, we preserve the
temporal structure during filtering.

\begin{figure*}  \fontsize{11}{20}\selectfont
\includegraphics[scale=1.14]{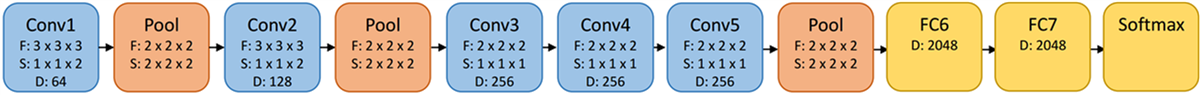} \caption{3D CNN architecture: the blue, orange and yellow blocks represent convolutional, pooling and fully-connected layers; \textit{F} is the filter size for the convolutional layers (or window size for pooling), formatted as width$\times$ height $\times$ time; \textit{S} is the filter/window stride;   \textit{D}\ is the number of filters (or number of hidden units) for the convolutional  and fully-connected layers. \label{fig:3D_CNN}} \end{figure*}

\subsubsection{Spatial Stream}
   Previous work has shown that stacking RGB frames channel-wise and ingesting such volumes into a 2D CNN does not necessarily improve performance  \cite{simonyan2014two, karpathy2014large}.  Indeed, one option is to train a deep 3D CNN on a 4D RGB frame input, which is the  proposed configuration for our temporal stream (with MV inputs). Whilst this has been  shown to improve performance with RGB frames \cite{tran2015learning}, it   is far more computationally expensive to implement when the inputs are at pixel resolution, i.e., typically $224 \times 224 $ for CNNs trained on ImageNet \cite{deng2009imagenet}. Therefore, the complexity of the network in terms of activations and weights quickly becomes unmanageable.   

Our approach alleviates these problems by simply ingesting single  RGB\ frames from the video as inputs to a 2D CNN, in order to exclusively model the  scene  semantics in the image; these comprise geometry, color and background information that can not be extracted from the P-frames directly.  For example, in the case of action recognition, the green grass and net and racket texture patterns in the frame could distinguish a sequence as being related to tennis, rather than swimming.   Given that such spatial structures and color information tends to be persistent  across frames belonging to the same type of scene, we can gain substantial storage and complexity savings by our proposed sparse full-frame decoding  and selective superpositioning of MB RGB texture decoding according to the motion activity, as described in Section \ref{sec:motion_estimation} and illustrated in  Fig. \ref{fig:mvx_seldec} and Fig. \ref{fig:activity_region}. 

In order to make our input independent of the video resolution, we follow the approach of Simonyan \textit{et al.} \cite{simonyan2014two}. That is, we first resize the RGB frame, such that the smaller side is equal to 256 and we keep the aspect ratio. From the resized frame $\boldsymbol{\Phi}_\text{S} \in
\mathbb{R}^{W_\text{S} \times H_\text{S} \times K_\text{S}}$, we crop/resize a fixed spatial size $N_\text{S} \times N_\text{S}$; $N_\text{S} = 224$. Our spatial stream input is thus of size $224 \times 224 \times 3.$

\subsection{Network Architecture} \label{sec:network_arch}

 Our 3D CNN architecture is illustrated in Fig. \ref{fig:3D_CNN}. All convolutions and pooling  are spatiotemporal in their extent. 3D pooling is
 performed over a $2 \times 2 \times 2$ window with spatiotemporal stride of 2.
 The first two convolutional layers use 3D filters of size $3 \times 3 \times 3$
 to learn spatiotemporal features. With a $24 \times 24 \times 2 \times 160$
 motion vector input, the third convolutional layer receives input of size
 $6 \times 6 \times 2 \times 10$.   Therefore, we set the filter size of the
 third, fourth and fifth convolutional layers to $2 \times 2 \times 2$, as this is sufficiently large to encompass the spatial extent of the
 input  over
 the three layers whilst minimizing the number of parameters.  In order to maintain
 efficiency when training/evaluating, we also use a temporal stride of  2 in the
 first and second convolutional layers to quickly downsize the motion vector
 input; in all other cases we set the stride to 1 for convolutional layers.  The temporal downsizing substantially minimizes the number of activations (and thus, the number of floating point operations) in the lower layers.  All
 convolutional layers and the FC6 \&\ FC7 layers use the
 parametric ReLU activation function \cite{he2015delving}.  

It is important to note that our network has substantially less parameters and activations than other
architectures using optical flow.  In particular,  our 3D CNN stores 29.4 million weights. For comparison, ClarifaiNet
\cite{zeiler2014visualizing} and similar configurations that are  commonly used for optical-flow based classification  \cite{simonyan2014two, zhang2016real} require roughly
100 million   parameters.

For the spatial stream, we opt for the commonly used VGG-16 \cite{simonyan2014very} architecture, as it is sufficiently deep to learn complex representations from the input frames. The CNN is typically trained on ImageNet \cite{deng2009imagenet} for image classification. While we have also obtained similar results with shallower networks,  VGG-16 allows for better generalization to larger datasets.

\subsection{Network Training} \label{sec:training_proc}
{We train on the temporal and spatial streams independently, as this permits sequential training on a single GPU and simplifies management of resources such as GPU RAM.  It additionally permits evaluation on a single stream for faster runtime.  The training details for each stream are as follows. }
\subsubsection{Temporal Stream}
We train the temporal stream using stochastic gradient descent
with momentum set to 0.9. The initialization of He \textit{et al.}
\cite{he2015delving}   is extended to 3D and the network weights are initialized from a normal
distribution with variance inversely proportional to the fan-in of the filter
inputs. Mini-batches of size 64 are generated by randomly selecting 64 training
videos.  From each of these  training videos, we choose a random index from
which to start extracting the P-frame MB motion vectors.  From this position, we
simply loop over the P-type MBs in temporal order    until we extract
motion vectors over $T$ consecutive P frames. This addresses the issue of videos
having less than $T$ total P frames, e.g., cases where the video is only a few
seconds long. For UCF-101, we train from scratch; the learning rate is initially
set to $10^{-2}$ and is decreased by a factor of $0.1$ every 30k
iterations. The training is completed after 70k iterations. Conversely, for HMDB-51,
we compensate for the small training split by initializing the network with  pre-trained weights from UCF-101 (split 1). The learning rate is initialized at $
10^{-3}$ and decayed by a factor of $0.1$ every 15k iterations,\ for 30k iterations. 

To minimize the chance of
overfitting due to the low spatial resolution    of these motion vector frames and the small size of the training split for both UCF-101 and HMDB-51, we
supplement the training with heavy data augmentation. To this end, we concatenate  the motion vectors into a single $W_\text{T} \times H_\text{T} \times
2T$ volume and apply the following steps; \textit{(i)}  a multi-scale random cropping  to fixed
size  $N_\text{c} \times N_\text{c} \times 2T$ from this volume, by randomly selecting a value
for $N_\text{c}$ from $N_\text{T} \times c$ with $c\in\{0.5, 0.667, 0.833, 1.0\}$; as such, the cropped
volume is randomly  flipped and spatially resized to $N_\text{T} \times N_{ \text{T}} \times 2T$; \textit{(ii)} zero-centering the volume by subtracting the  mean motion vector value from each motion
vector field $\boldsymbol{\Phi}_\text{T}$,  in order to remove possible bias; the
$\delta x$ and $\delta y$ motion vector components can now be split into
separate channels,  thus generating our 4D network input
$\boldsymbol{\hat{\Phi}}_\text{T}$. During training, we additionally regularize the
network by using dropout ratio of  0.8 on the FC6 and FC7 layers together with weight decay of 0.005.

\subsubsection{Spatial Stream}

We also train the spatial stream independently using stochastic gradient descent with momentum set to 0.9.  As with the temporal stream, mini-batches of 64 are amalgamated over 64 randomly selected videos. We take advantage of the transferability of features from  image to video classification, and pretrain all layers of our VGG-16 architecture on ILSVRC'12 \cite{krizhevsky2012imagenet}; all layers are subsequently fine-tuned on the video training sets.  The learning rate is initialized at $10^{-3}$ and decayed by a factor of $0.1.$ We complete training at 15k iterations. 

Again, due to the small training sizes, we risk overfitting during training; therefore we set dropout and weight decay on the first two fully connected layers to 0.8 and 0.005 respectively.
We also use a multi-scale random cropping of the resized RGB frame by randomly selecting a value from $N_\text{S} \times d$ with $d \in \{0.857, 1.0,  1.143\}$; the cropped volume is subsequently randomly flipped, spatially resized to $N_\text{S} \times N_\text{S} \times 3$ and zero-centered as per the temporal stream.  
\subsection{Testing}\label{sec:Testing} 

During testing, per video, we  generate 2  
volumes  of temporal size $T$ from which to evaluate on the temporal stream. The starting indices for the volumes are at the first P-frame and  at half the total number of P-frames.   Per volume, we crop the four corners,    the center of
the image (and its mirror image) to size $N_\text{T} \times N_\text{T} \times 2 \times T$.  In order to generate our prediction for the video, we take the maximum score over all crops. Due to the low resolution and short duration of the HMDB-51 and UCF-101 videos, taking these extra crops and volumes is often redundant as the spatial resolution of the P-frames is  low and the temporal extent $T$ of the input is large enough to encompass the entire video duration.  However, our approach is better suited to videos ``in the wild" and  we can afford the use of extra crops due to the low complexity of our 3D CNN.

We evaluate  on the spatial stream by extracting only 5 frames from the set per video, albeit with only a  single center crop (and its horizontal flip) of size $N_\text{S} \times N_\text{S} \times 3$.  In our experiments, we have found this to be sufficient for the case of trimmed action recognition, where most frames are relevant to the associated video label. The frames are extracted at evenly spaced intervals from the video. To generate our prediction, we again compute the maximum score over all extracted frames. In order to produce a final score for the fusion of the two modalities, we simply average their maximum scores, which is equivalent to combining knowledge from the most relevant input in each stream.  

\section{Experimental Results}\label{sec:Evaluation}


\subsection{End-Point Error and Speed of MB MV Extraction and Decoding vs. Optical Flow Methods used in Video Classification}



In order to examine the accuracy and extraction time of our approach versus decoding and optical flow estimation, we perform a comparison against the Brox \cite{brox2011large} and
FlowNet2 \cite{ilg2016flownet} optical flow estimation methods that were respectively used (amongst others) by Simonyan \textit{et al.} \cite{simonyan2014two} and Brox \textit{et al.} \cite{ilg2016flownet}. Table \ref{tab:codec_stat} presents the motion field estimation accuracy, measured in terms of end-point error (EPE) on MPI-Sintel,  for which ground truth motion flow is also available (see Fig. \ref{fig:compmv}). Since our CNN architecture downsamples the optical flow before ingestion \cite{simonyan2014two}, we measure the EPE  for our MV flow estimation at the resolution of our CNN input.      Under these settings,  Table \ref{tab:codec_stat}   shows that the EPE of our approach is 1.75 to 4.86 times higher than that of optical flow methods. Despite the detrimental accuracy, our EPE results remain low enough to
indicate high correlation with the ground-truth motion flow and the optical-flow based methods. Indeed, the results  presented in the following subsections show that the codec MB MV accuracy suffices for classification results that are competitive to the state of the art.

In order to measure flow estimation and  decoding speed (with I/O) in terms of frames per second (FPS), we now use video content that corresponds to our video classification tests, i.e., 100 video sequences from UCF-101 (see next subsection for the details of this dataset). All CPU-based experiments were carried out on an Amazon Web Services (AWS)\ EC2 r3.xlarge instance (Intel Xeon E5-2670 v2 CPU), while all GPU-based experiments were carried out on a AWS EC2 p2.xlarge instance (Tesla K80 GPU). For our selective decoding approach described in Section \ref{sec:motion_estimation}, we select values for the decoding interval $X$ that correspond to the settings used in our video classification tests. The results of this experiment are summarised in Table \ref{tab:decode_stat}. In terms of  flow estimation speed, our CPU-based MV flow extraction is more than 1500 times faster than FlowNet2 and more than 977 times faster than Brox flow (both running on a GPU), as it does not require video decoding or any optical flow computation.  At current AWS pricing\footnote{\href{https://aws.amazon.com/ec2/pricing/on-demand/}{AWS EC2 spot pricing}, (r3.xlarge vs. p2.xlarge N. Virginia, Sept. 2017)}, GPU instances require more than 2.7 times the cost of CPU instances; as such, our AWS-based implementation has more than 2600 times lower cost.  This means that, for the public cloud cost that an optical flow method will process 1 hour of video, our approach will be able to process more than three and a half months of video footage.

\begin{table}[t] \caption{Motion field end-point error (EPE) for the proposed approach, Brox  \cite{brox2011large} and FlowNet2  \cite{ilg2016flownet}.   } \centering \begin{footnotesize} \resizebox{100pt}{!}{\begin{tabular}{cccc} \toprule
                Input    & EPE \\  \midrule
Proposed, MV      & 15.26 \\

Brox                      & 8.70 \\

FlowNet2                    & 3.14 \\  \bottomrule

 \end{tabular}} \end{footnotesize}   \label{tab:codec_stat} \end{table}

\begin{table}[t] \caption{Flow estimation and  decoding speed results for the proposed approach, Brox  \cite{brox2011large} and FlowNet2  \cite{ilg2016flownet}.} \centering \begin{footnotesize} \resizebox{210pt}{!}{\begin{tabular}{cccc} \toprule
                \multirow{2}{*}{Input}  & \multicolumn{2}{c}{Frames Per Second (FPS)} \\
                &Flow Estimation &Decoding 

\\  \midrule
Proposed,  $X=10$ & 18226 (CPU) & 1180 (CPU)   \\

Proposed,  $X=50$ & 18226 (CPU) &2016 (CPU)   

 \\

Brox                  & 18.64 (GPU) & 168 (CPU)  \\

FlowNet2               & 12.08 (GPU) & 168 (CPU)  \\  \bottomrule

\end{tabular}} \end{footnotesize}   \label{tab:decode_stat} \end{table}

\begin{figure}[t!] \centering \includegraphics[scale=0.25]{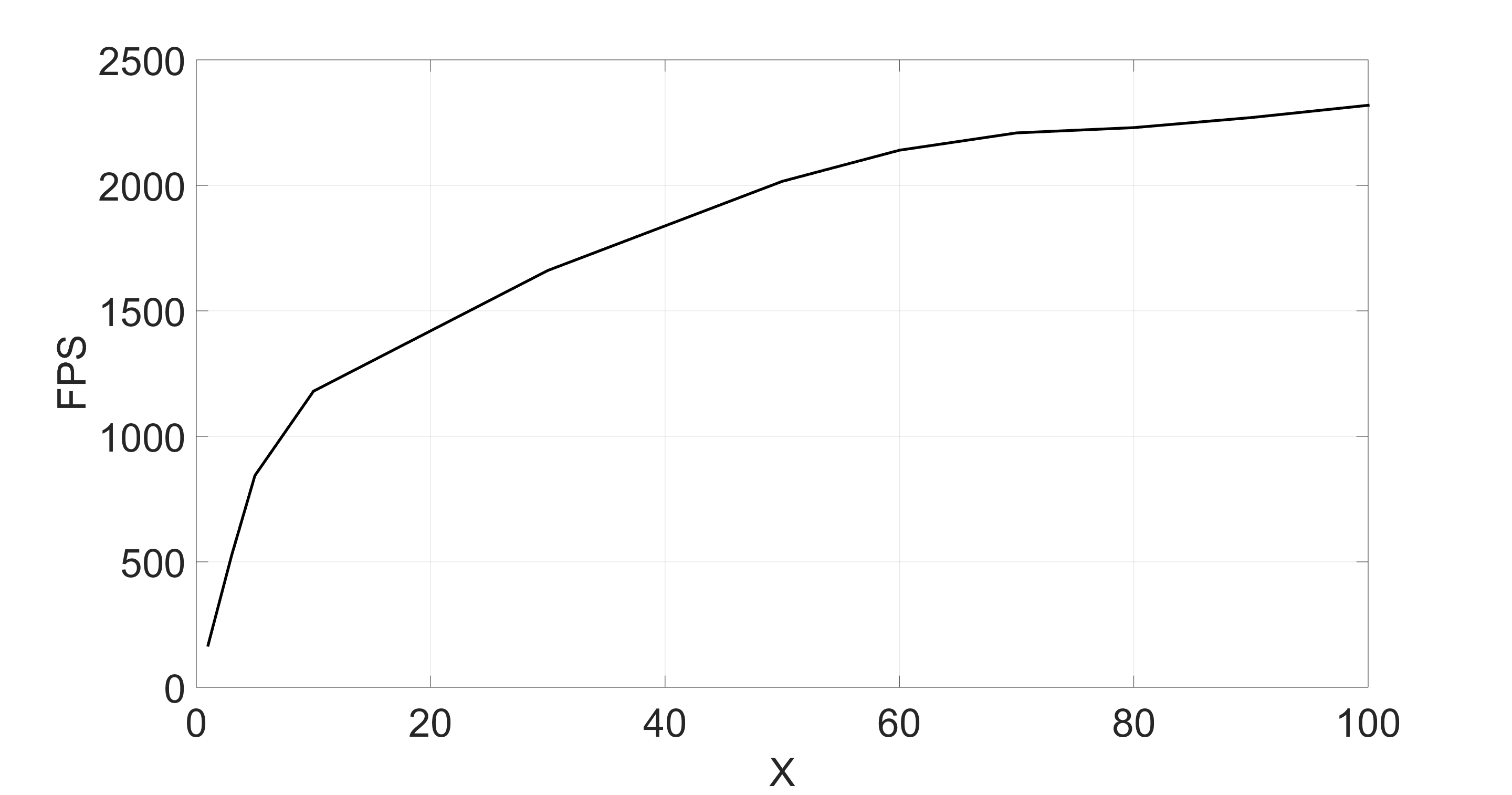}
       \caption{\label{fig:xfps} Achieved FPS of selective decoding for varying decoding interval $X$. } \end{figure}

\begin{figure}[t!] \centering \includegraphics[scale=0.25]{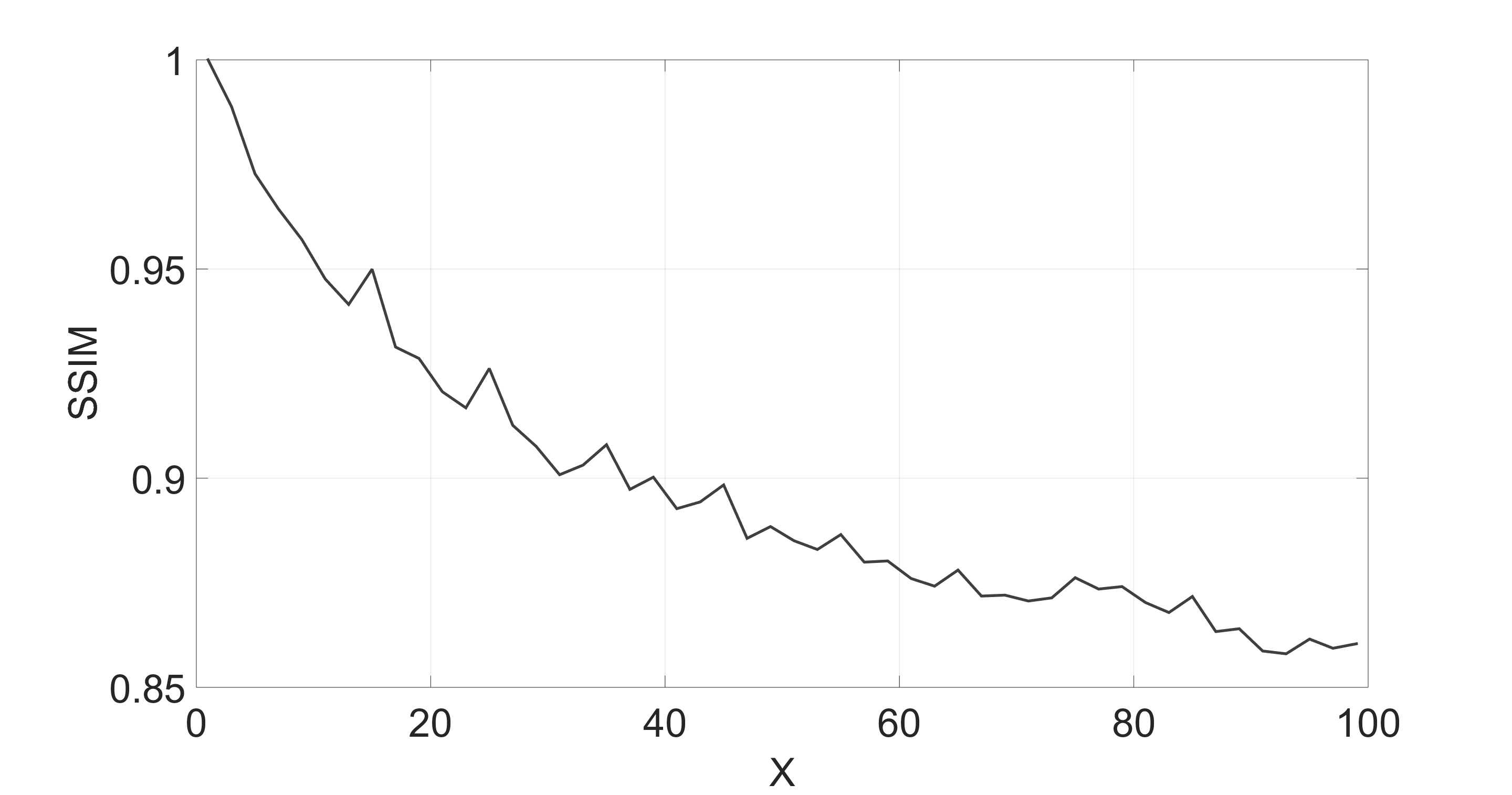}
       \caption{\label{fig:ssim} Structural similarity index metric (SSIM) for varying decoding interval $X$.} \end{figure}

In terms of  decoding speed, Table \ref{tab:decode_stat}  shows that selective decoding is  an order of magnitude faster than the full-frame decoding required for Brox and FlowNet2. We illustrate the influence of selective decoding on the achieved FPS in more detail in      Fig. \ref{fig:xfps}. The results show that the decoding FPS increases rapidly until $X >50$ and begins to saturate after this point.    In order to associate this speed up with a measure for the expected   visual quality of the selective decoding  and rendering approach,  we plot the average structural similarity index (SSIM) \cite{wang2004image} for multiple values of  $X$ in Fig. \ref{fig:ssim}, using the fully-decoded video sequences as reference.    By combining the two figures, it is evident that, as  the decoding speed increases and reaches a saturation at around 2500 FPS,  the quality of all rendered frames decreases and plateaus at  SSIM values around 0.85. We next assess whether the motion flow accuracy and visual quality allow for high-performant video classification with the proposed CNN-based architectures.

\subsection{Datasets used for Video Classification} \label{sec:datasets}

Evaluation is performed on two standard action recognition datasets, UCF-101
\cite{soomro2012ucf101} and HMDB-51 \cite{{kuehne2011hmdb}}.  UCF-101 is a
popular action recognition dataset, comprising 13K videos from 101 action
categories with   $320 \times 240$ pixels per frame, at replay rate of 25 frames per second (FPS).  HMDB-51 is a
considerably smaller dataset, comprising only 7K videos from 51 action
categories, with the same spatial resolution as UCF-101, and at 30 FPS replay rate.
{Finally, unless  stated otherwise, we always cross-validate on the standard three splits for both datasets.}

\subsection{Evaluation Protocol and Results} \label{sec:eval_pro}

For each dataset we follow the testing protocol of Section \ref{sec:Testing}.  Each UCF-101 training split
consists of approximately 9.5K videos, whereas each HMDB training split has 3.7K videos. We report all single stream  feedforward network runtimes without I/O, in order to isolate the efficiency of our proposed architecture. Speed is reported in terms of FPS, which is computed as  the number of  videos each network can process per second multiplied by the average number of frames per video (we use the average length of UCF-101 videos, i.e.,  180 frames \cite{simonyan2014two}). By using FPS as our metric, we account for both  the network complexity and  the   number of inputs processed per video at inference, i.e., the number of crops and volumes taken, as reported in the respective papers.  For frameworks where the number of inputs is a function of the video size, we again assume an average video length of 180 frames.  All speed results correspond to a batch size of 32 on  an AWS\ EC2 p2.xlarge instance, which comprises a single K80 GPU.

\begin{table}[ht] 
\caption{Classification accuracy and speed (FPS) against state-of-the-art flow based networks. ``Proposed 3D CNN'' refers to our  temporal stream that ingests MB motion vectors.  }
\centering
\resizebox{\linewidth}{!}{
\begin{tabular}{ccccccc}
 
\toprule 

Framework & Input  &
\multicolumn{2}{c}{Accuracy (\%)} & FPS  \\

& Size   &  \begin{scriptsize}UCF\end{scriptsize} & \begin{scriptsize}HMDB\end{scriptsize}  \\ \midrule

Proposed 3D CNN   & 24$^{2}\times$2$\times$160  & 77.2 & 48.0 &  3105  \\

TSCNN-Brox \cite{simonyan2014two}  & 224$^2 \times$20  & 81.2 & 55.4 &185  \\

LTC-Brox \cite{varol2016long} & 58$^2\times$2$\times$100    & 82.6 & 56.7 & \textless100\ \\
LTC-Mpegflow \cite{varol2016long}& 58$^2\times$2$\times$60  & 63.8 & -- & \textless100\\

TSCNN-FlowNet2 \cite{ilg2016flownet}  &224$^2 \times$20  & 79.5 & -- &185\\

EMV-CNN (ST+TI) \cite{zhang2016real} & 224$^2 \times$ 20  & 79.3 & --  & 1537\\

\bottomrule 
\end{tabular}}
\label{tab:temporal_results}
\end{table}

\begin{table}[t] 
\caption{Complexity of proposed 3D CNN vs EMV-CNN with respect to millions of activations and weights (\#A, \#W), summed over conv, pool and FC layers in the utilized deep CNN of each approach. }
\centering \resizebox{170pt}{!}{\begin{tabular}{cccc}
 
\toprule 

Framework   &
\multicolumn{2}{c}{Complexity} \\

  &  \begin{scriptsize}\#A($\times 10^6$)\end{scriptsize} &\begin{scriptsize}\#W($\times 10^6$)\end{scriptsize}   \\ \midrule

Proposed 3D CNN   & 4.0 & 29.4   \\

EMV-CNN \cite{zhang2016real} & 2.0 & 90.6 \\

\bottomrule 
\end{tabular}} 
 \label{tab:network_complexity}
\end{table}

\subsubsection{Temporal Stream}

Table \ref{tab:temporal_results} presents the results of temporal stream CNNs on split 1 of the datasets, for which our method achieves 77.2\% and 48.0\% on UCF-101 and HMDB-51, respectively.  When cross-validating on all three splits for both datasets, our accuracy is higher and we achieve 77.5\%\ on UCF-101 and 49.5\%\ on HMDB-51.  It is evident that our approach performs competitively to recent proposals utilizing highly-complex optical flow, whilst minimizing the network complexity via the low number of activations in the lower convolutional layers and small spatial size of the input. As a consequence of the lower resolution inputs and longer temporal extents, our proposal is able to achieve 2 to 30-fold higher FPS in comparison to all other frameworks\footnote{We remark that 
 Laptev \textit{et al.}  \cite{mpegflow}  made a proposal that uses codec MVs; their method is based on the
encoding of such MVs into  Fisher vectors (instead of CNNs)\ to classify video activity.  However, that approach is only capable of  achieving an accuracy
of 46.7\%\ on HMDB \cite{mpegflow} at a much lower frame rate (130 FPS) compared to recent CNN methods.}. 


The closest competitor is the MV based EMV-CNN method \cite{zhang2016real}, which achieves approximately half the FPS of our approach and therefore warrants further discussion.  During test-time,   
 EMV-CNN stacks 10 P or B frames as input to their temporal stream, whereas we
stack 160 P-frames per input to our 3D CNN. {According to  the GOP structure, for every P-frame, EMV-CNN must additionally extract and process 2 B-frames, whereas our 160 P-frame temporal extent   typically constitutes processing the entire video
in one forward-pass. As such, we  only require 12 inputs (2 volumes, 6 crops per volume) to classify a video from UCF-101, whereas EMV-CNN evaluates on 25 inputs per video. Importantly, unlike our approach, EMV-CNN requires
optical-flow based training with teacher initialization and supervision transfer from the optical flow based training \cite{zhang2016real}. This has the following detriments: 
\begin{itemize}
\item It makes any gains stemming from codec MVs negligible
during training and limits the scale-up of training. 
\item It requires upsampled P and B-frame MV fields due to the supervision transfer, which leads to reduced MV extraction and\ CNN processing speed in comparison to our proposal. 
\end{itemize}} 
In order to go into more detail on the complexity of our CNN against the one proposed within EMV-CNN, we present their network complexity in Table \ref{tab:network_complexity}. The EMV-CNN architecture requires (approximately) 3 times the number of weights (and thus 3 times the memory) of our 3D\ CNN. 

Finally, as discussed in Section \ref{sec:Testing},  due to the low resolution of our input, taking a large number of crops  is redundant in our proposal.  Therefore, it should be  possible to achieve similar performance even in the case of with one-shot recognition.  Indeed, when evaluating our temporal stream on a single center crop, we achieve 76.2\% on UCF-101 (split 1) and 46.7\% on HMDB-51 (split 1). Such a simplification increases the frame rate to 7452 FPS.


\subsubsection{Spatial Stream} With regards to the spatial RGB\ stream produced by the proposed selective decoding,  Table \ref{tab:spatial_results} presents results  with two values of $X$. The network is evaluated and accuracy is subsequently averaged over cross-validation with the  three splits for both datasets. The first result ($X=10$) fully decodes  every 10 frames, whilst the second result corresponds to selective decoding and rendering every 10 frames and full decoding every 50 frames.  In the latter case, we set the rendering frame interval to  $R=10$ and threshold $A=0$, i.e., selectively decoding and writing the RGB\ texture of macroblocks corresponding to non-zero motion vectors once every 10 frames. As the number of crops/volumes that the network evaluates on is fixed and independent of the fully decoding interval  $X$ and rendering frame interval $R$, the network FPS reported in Table \ref{tab:spatial_results} is the same for both cases. However, the decoding runtime for $X =50$ is  approximately 1.7 times faster than for $X=10$ (see Table \ref{tab:decode_stat}).
The results demonstrate that the performance drop from  selective decoding is marginal and that our spatial stream proposal significantly outperforms TSCNN \cite{simonyan2014two} and SFCNN \cite{karpathy2014large} on UCF-101, whilst performing inference with approximately 5 times higher speed. We achieve this speed by restricting the inputs to single frames and only evaluating on 10 inputs per video, which counterbalances the higher complexity of our pretrained VGG16 network. On the contrary, approaches like LTC \cite{varol2016long}, TSCNN \cite{simonyan2014two} and C3D\cite{tran2015learning} evaluate on many more frames and multiple crops per frame, thereby   incurring higher computational overhead and significantly lower frame rate for their evaluation process.  However, it is worth noting that when comparing our proposed temporal and spatial stream FPS, the temporal stream runs approximately 2.5 times faster, which further motivates training the streams independently, as  we can easily allocate more resources to processing the slower stream.

\begin{table}[t] 
\caption{Classification accuracy and runtime (FPS) against state-of-the-art RGB based networks.   For our proposed spatial streams, we fully decode one frame every $X$ frames.  }
\centering
\resizebox{\linewidth}{!}{
\begin{tabular}{ccccccc}
 
\toprule 

Framework & Input  &
\multicolumn{2}{c}{Accuracy (\%)} & FPS \\

& Size  &   \begin{scriptsize}UCF\end{scriptsize} & \begin{scriptsize}HMDB\end{scriptsize}   \\\midrule

 Proposed, $ X=10$ & 224$^{2}\times$3 & 79.3 &  42.4  & 1228\\
 
 Proposed, $X=50$ & 224$^{2}\times$3 & 77.7 & 39.6 & 1228 \\

TSCNN \cite{simonyan2014two} &  224$^{2}\times$3  & 73.0 & 40.5 & 252 \\

SFCNN \cite{karpathy2014large} & 170$^2\times$3$\times$10 & 65.4 & -- & 216 \\

LTC \cite{varol2016long} & 71$^2 \times$3$\times$100& 82.4 & -- &  \textless100  \\

C3D\cite{tran2015learning}  & 112$^2\times$3$\times$16   & 82.3 & -- & \textless300  \\

\bottomrule 
\end{tabular}}
\label{tab:spatial_results}
\end{table}

\subsubsection{Spatio-temporal stream fusion and complementarity in predictions}

\begin{table}[ht] 
\caption{Comparison against state-of-the-art fusion based frameworks.   For our proposed two stream networks, the spatial stream ingests one fully-decoded RGB frame every $X$ frames. }
\centering
\resizebox{200pt}{!}{
\begin{tabular}{cccccc}
 
\toprule 

Framework & 
\multicolumn{2}{c}{Accuracy (\%)} &\ \\

&  \begin{scriptsize}UCF\end{scriptsize} & \begin{scriptsize}HMDB\end{scriptsize} \\ \midrule

 Proposed, $X=10$    & 89.8 & 56.0   \\
 
 Proposed, $X=50$   & 88.9 & 54.6 \\

TSCNN (avg. fusion) \cite{simonyan2014two} &  86.9 & 58.0\\

TSCNN (SVM fusion) \cite{simonyan2014two} &  88.0 & 59.4 \\

CNN-pool \cite{yue2015beyond} &  88.2 & -- \\

C3D (3 nets)+IDT\cite{tran2015learning}  &  90.4 & -- \\

LTC\cite{varol2016long}  &  91.7 & 64.8\ \\

EMV + RGB-CNN \cite{zhang2016real} & 86.4 & --\\

IP+SVM \cite{xu2017two} & -- & 59.5 \\

Line Pooling \cite{zhao2017pooling} & 88.9 & 62.2 \\

TDD \cite{wang2015action} & 90.3 & 63.2 \\
 
\bottomrule 
\end{tabular}}
 \label{tab:fusion_results}
\end{table}

Table \ref{tab:fusion_results} presents the summary of the classification performance of our proposed two-stream approach (averaged over the standard three splits) when fusing the spatial and temporal streams. Our two-stream network utilizing selective decoding achieves 89.8\% accuracy on UCF-101. Overall, our approach is within a few percentile points from the best results reported for both datasets, whilst skipping the complex preprocessing inherent with decoding and optical flow based methods. TSCNN, LTC and Line Pooling all use Brox optical flow in their temporal streams, whilst IP+SVM uses a combination of optical flow, pixel values and gradients for descriptor computation. On the other hand, methods such as C3D+IDT and Line Pooling use trajectory-based descriptor computation in their fusion based frameworks, in addition to deep CNN computation,  which   adds even further complexity to the classification pipeline. Line Pooling also adopts VGG-16 in their spatial stream as in our case, but forgoes our simpler end-to-end approach for frame pooling and VLAD\ encoding on an intermediate convolutional layer, which requires additional codebook learning.

Given that our spatio-temporal stream fusion strategy increases the accuracy by  8\%-12\%\ in comparison to independent stream evaluation, further investigation of the inference properties of the spatial and temporal CNNs is warranted. The temporal stream ingests inputs of low spatial resolution with long temporal extent, whereas the spatial stream ingests inputs with high spatial resolution but low temporal resolution (single frames). As such, the temporal and spatial stream  raters are expected to be more disjoint in their learned representations, which translates to higher information gain and a  significant increase in accuracy when their inferences are fused.  To quantify their pairwise agreement, we compute Cohen's  kappa   $\kappa$  \cite{cohen1960coefficient} between the raters producing labels for the: temporal stream, spatial stream, two-stream and the  ground truth (i.e., ``null''\ model) 

\begin{figure}[t!] \centering \includegraphics[scale=0.25]{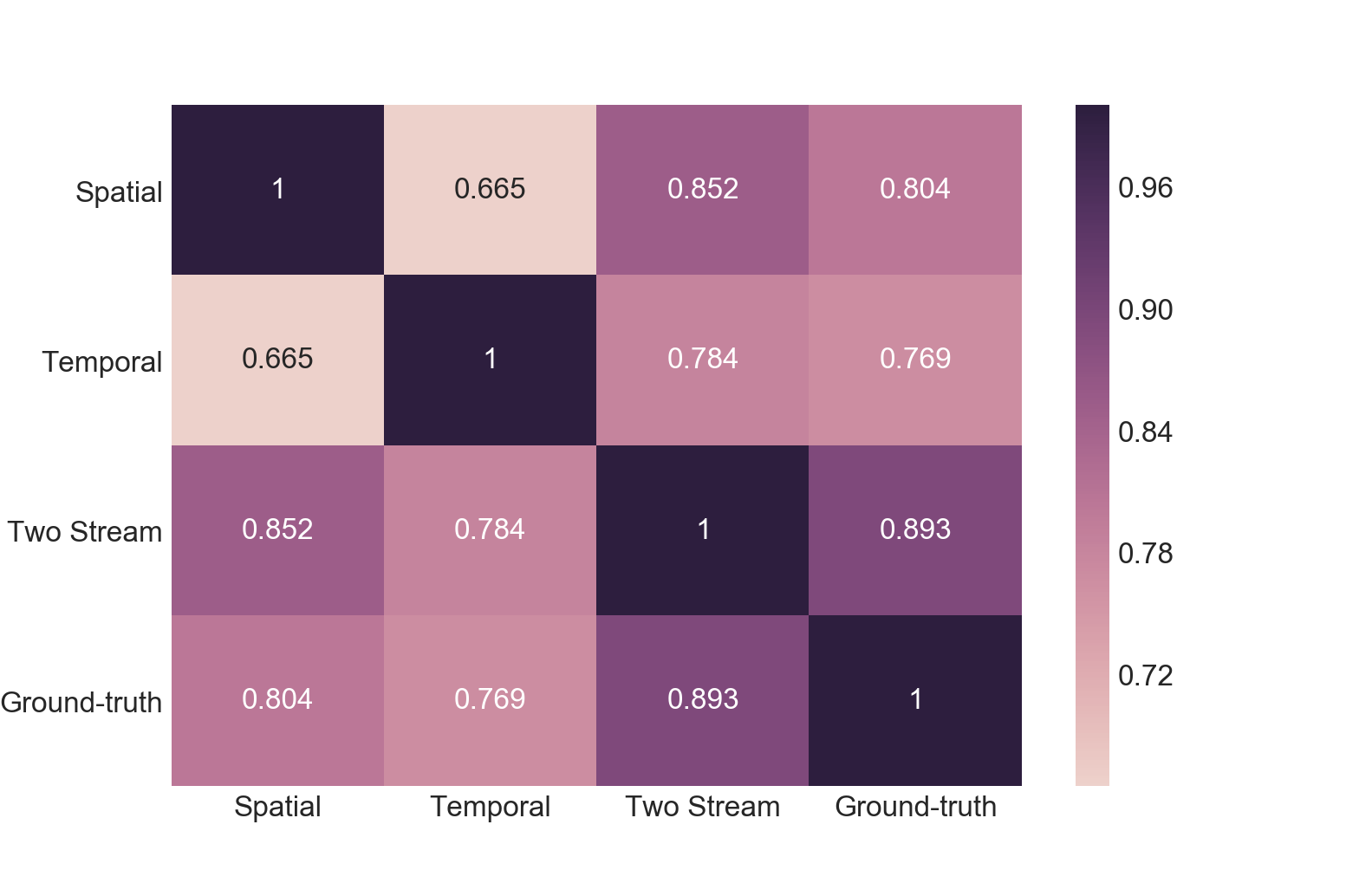}
       \caption{\label{fig:kappa_matrix} Cohen's kappa matrix over all rater combinations for UCF-101 (split 1). For the spatial stream,  $X = 10, R = 10$.} 
\end{figure}

\begin{equation}
\kappa = \frac{p_o - p_e}{1-p_e}
\end{equation}
where $p_o$ is the relative observed agreement amongst raters (equivalent to accuracy) and $p_e$ is a hypothetical probability of random agreement, which is summation  of marginal probabilities multiplied between raters. Cohen's kappa ranges from $\kappa = 1.0$ (raters of in complete agreement) to
$\kappa \leq 0$ (no agreement amongst raters other than random expectation). We plot the symmetric matrix of $\kappa$ values over all rater combinations for predictions on UCF-101 (split 1) in Fig. \ref{fig:kappa_matrix}. \ As expected, there is high inter-rater agreement between the spatio-temporal two-stream architecture and the ground-truth, with $\kappa(\text{two-stream, ground-truth}) = 0.893$.  However, there is low inter-rater agreement between the independent spatial and temporal streams, i.e.,  $\kappa(\text{spatial, temporal}) = 0.665$. This shows that the temporal and spatial streams learn  heterogenous representations that reliably classify different video subsets from the dataset.  Indeed, if these video subsets were very similar to each other, then $\kappa(\text{spatial, temporal})$ would be approximately equal to $\min(\kappa(\text{spatial, ground-truth}), \kappa(\text{temporal, ground-truth}))=0.769$; however, this is approximately 10\% higher than the actual value of $\kappa(\text{spatial,temporal})$ and also turns out to be the average gain obtained by the spatio-temporal fusion. 

\begin{figure}[t!] \centering \includegraphics[scale=0.3]{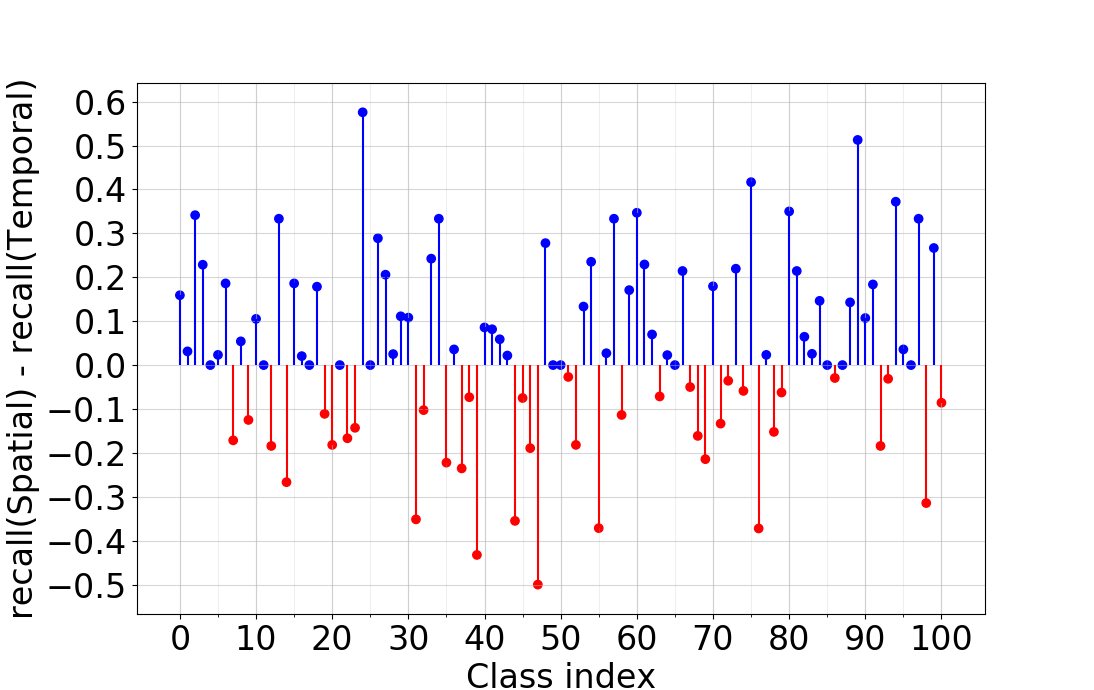}
       \caption{\label{fig:recall_ucf} Recall difference graph for UCF-101 (split 1).  Red line equates to a temporal bias, blue line equates to spatial bias.  Classes are in alphabetical order.} \end{figure}

\begin{figure}[t!] \centering \includegraphics[scale=0.3]{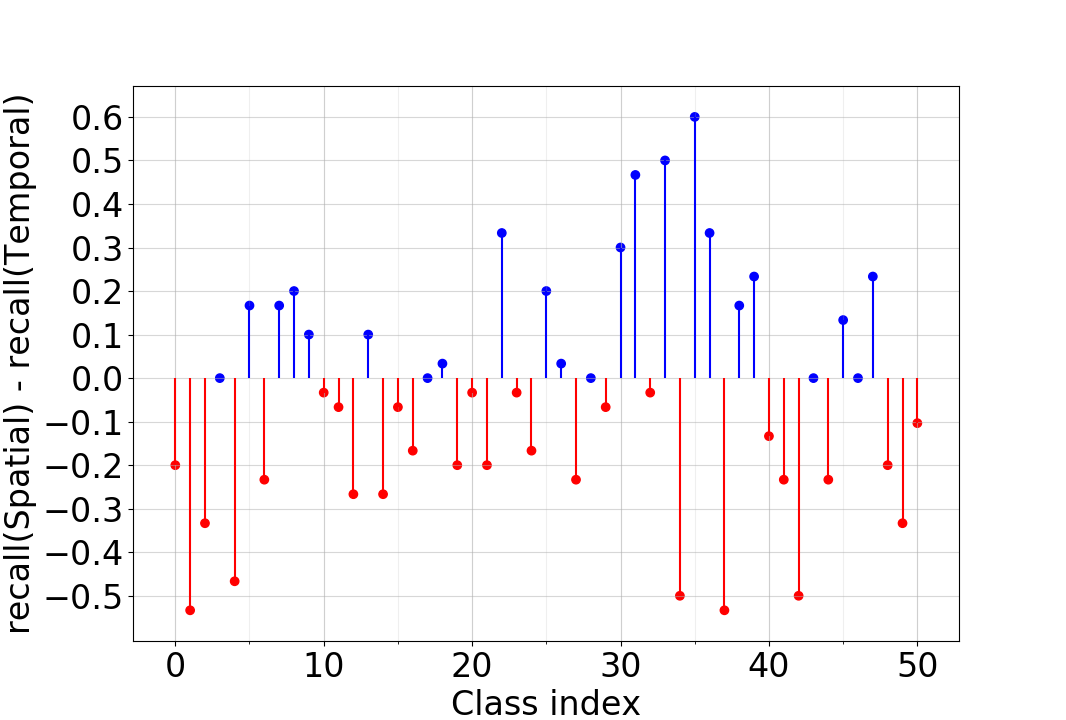}
       \caption{\label{fig:recall_hmdb} Recall difference graph for HMDB-51 (split 1). Red line equates to a temporal bias, blue line equates to spatial bias.  Classes are in alphabetical order.} \end{figure}

In order to reinforce this point, we measure the difference in recall values for each class between the spatial and temporal stream, and plot this in Fig. \ref{fig:recall_ucf} and Fig. \ref{fig:recall_hmdb} for UCF-101 and HMDB-51 respectively.  If the two streams are in complete agreement, one would expect the recall difference to be close to 0; in other words, the two streams agree on the same video subsets. However, what we observe is that the temporal and spatial streams exhibit distinct  biases  in terms of class, depending on the nature of the activity.   For UCF-101, the temporal stream is favorable for ``high activity'' classes, such as ``high jump'' (class 39), ``jump rope'' (class 47) and ``salsa spin'' (class 76).  Conversely, the spatial stream  performs better for ``low activity'' classes where a scene representation is more informative, such as ``cutting in kitchen'' (class 24), ``rowing'' (class 75) and ``table tennis shot'' (class 89).  Intuitively, this means the stream fusion provides better generalization over all classes, or a lower network variance.

\begin{table*}[t] \caption{Cost per component and  end-to-end cost $C_{\text{tot}}$ for our proposed two-stream framework versus competitive frameworks to perform inference on UCF-101 (split 1).   } \centering \resizebox{350pt}{!} {\begin{tabular}{cccccc} \toprule
                Framework    & $C_{\text{flow}}$ (\$) & $C_{\text{decode}}$ (\$) & $C_{\text{t-stream}} (\$)$ &$C_{\text{s-stream}}$ (\$) & $C_{\text{tot}}$  (\$) \\  \midrule
Proposed, $X=10$      & 0.003 & 0.053 & 0.055 & 0.139 & 0.250\\

Proposed, $X=50$  & 0.003 & 0.031  & 0.055 & 0.139 & 0.228  \\

TSCNN (fusion) \cite{simonyan2014two} &  9.133  & 0.375  & 0.920 & 0.676  & 11.103  \\

EMV + RGB-CNN \cite{zhang2016real}  & 0.006 & 0.375 & 0.111 & 0.676 & 1.167 \\  \bottomrule

 \end{tabular}}    \label{tab:total_cost} \end{table*}

Finally, in order to associate our implementation results with the cost incurred by the two fastest methods of Table \ref{tab:fusion_results}, namely TSCNN (avg. fusion) and EMV + RGB-CNN, we present the AWS\ deployment cost of each method  in  Table \ref{tab:total_cost}. The table shows the cost incurred per component of each method,  as well as the total end-to-end cost, $C_{\text{tot}}$. The four components benchmarked in the table are: flow estimation, decoding, temporal stream inference and spatial stream inference, with costs: 

\begin{equation}
C_{\text{component}} = \frac{A}{3600}\times\frac{P_{\text{component}}}{F_{\text{component}}}, 
\label{cost_component_eq}
\end{equation}

where: $A$ is the average number of frames required for inference
in UCF-101  split 1 (180 frames/video $\times$ 3783 videos in split 1),   $F_\text{component}$ comprises the FPS results reported in Tables \ref{tab:codec_stat}-\ref{tab:spatial_results} for $\text{component}\in\{\text{flow, decode, t-stream, s-stream}\}$ of each method, and $P_\text{component}$ is the  \$/hr cost of the AWS instance  used to achieve the reported FPS.  Specifically, based on the on-demand cost for a p2.xlarge  (K80 GPU\ instance)\ and r3.xlarge (quadcore CPU)\ instance: 
\begin{itemize}
\item
 $P_\text{t-stream}=P_\text{s-stream}=0.9$  \$/hr and $P_\text{decode}=0.333$  \$/hr,   
\item
$P_\text{flow}=0.333$  \$/hr for our proposal and EMV-CNN,  
\item
$P_\text{flow}=0.9$  \$/hr for TSCNN because it requires GPU-based Brox flow estimation. \end{itemize}
The total cost $C_{\text{tot}}$  to process the UCF-101 (split 1) is:    

\begin{equation}
C_{\text{tot}} = C_{\text{flow}} + C_{\text{decode}} + C_{\text{t-stream}} + C_{\text{s-stream}}.
\label{cost_tot_eq}
\end{equation}

From Table \ref{tab:total_cost}, it is evident that the cost of dense optical flow in TSCNN completely overshadows all other costs. On the other hand, our MV flow  incurs less than 0.005\% its cost.  In addition, the combination of: 
\begin{itemize}
\item selective decoding (that incurs only 8.3\% the cost of full frame decoding with $X=50$), 
\item the more efficient CNN processing, and 
\item the cost $C_\text{flow}$ of our MV flow estimation being approximately  half that of EMV + RGB-CNN due to our temporal CNN only requiring $24\times 24$ crops of the P-frame MV field (while  EMV-CNN ingests upsampled P and B-frame MV fields to carry out the supervision transfer  \cite{zhang2016real}),
\end{itemize}
lead to the proposed method incurring only 20\% of the cost of  EMV + RGB-CNN for inference. Overall, our approach is found to be 5 to 49 times cheaper to deploy on AWS\ than the most efficient methods from the state-of-the-art in video classification.

\section{Conclusion}\label{sec:Conclusion}
We propose a 3D CNN architecture for video classification that utilizes compressed-domain motion vector information for substantial gains in speed and implementation cost on public cloud platforms. We fuse the �3D CNN  with a spatial stream that ingests selectively decoded frames, determined by the motion vector activity.  Our MV\ extraction is found to be three orders of magnitude faster than optical flow methods. In addition, the selective macroblock RGB\ decoding is one order of magnitude faster than full-frame decoding. By coupling the high MV extraction and\ selective RGB\ decoding speed  with lightweight CNN processing, we are able to classify videos with one to two orders of magnitude  lower cloud computing cost in comparison to the most efficient proposals from the literature, whilst maintaining competitive classification accuracy (Tables \ref{tab:fusion_results} and \ref{tab:total_cost}).   Further refinements of our approach may allow for the first time CNN-based classification of exascale-level video collections to take place via commodity hardware, something that currently remains unattainable by all CNN-based video classification methods that base their training on full-frame video decoding and optical flow estimation. Source code related to the proposed approach is available online, at \href{http://www.github.com/mvcnn}{http://www.github.com/mvcnn}.

{\small
\bibliographystyle{IEEEtran}
\bibliography{literature}
}

\end{document}